\documentclass[10pt,twocolumn,letterpaper]{article}

\usepackage{cvpr}              % To produce the CAMERA-READY version
\definecolor{cvprblue}{rgb}{0.21,0.49,0.74}
\usepackage[pagebackref,breaklinks,colorlinks,allcolors=cvprblue]{hyperref}

\def\paperID{6948} % *** Enter the Paper ID here
\def\confName{CVPR}
\def\confYear{2025}

\title{Categorical Keypoint Positional Embedding for Robust Animal Re-Identification}

\author{Yuhao Lin, Lingqiao Liu, Javen Qinfeng Shi\\
Australian Institution for Machine Learning\\
The University of Adelaide\\
{\tt\small yuhao.lin01,lingqiao.liu,javen.shi@adelaide.edu.au}
}

\begin{document}
\maketitle
\begin{abstract}
% Animal re-identification (ReID) has emerged as a critical technology for ecological research, crucial for monitoring population dynamics, analyzing behavioral patterns, and assessing ecological impacts vital for conservation strategies. Unlike human ReID, animal ReID is challenged by highly variable poses, diverse environmental conditions, and frequent occlusions that obscure key visual features. This study introduces a novel approach to animal ReID that utilizes a groundbreaking keypoint propagation technique, employing a single annotated image to propagate keypoints across an entire dataset via a pre-trained diffusion model. Additionally, we enhance the Vision Transformer's (ViT) capability to recognize and contextualize these keypoints through Keypoint Positional embedding (KPE) and Categorical Keypoint Positional Embedding (CKPE). Our results significantly outperform existing state-of-the-art methods in wildlife ReID tasks, with marked accuracy improvements on established benchmarks, demonstrating the efficacy of our approach in tackling the unique challenges of animal ReID. Code will be released.

Animal re-identification (ReID) has become an indispensable tool in ecological research, playing a critical role in tracking population dynamics, analyzing behavioral patterns, and assessing ecological impacts, all of which are vital for informed conservation strategies. Unlike human ReID, animal ReID faces significant challenges due to the high variability in animal poses, diverse environmental conditions, and the inability to directly apply pre-trained models to animal data, making the identification process across species more complex. This work introduces an innovative keypoint propagation mechanism, which utilizes a single annotated image and a pre-trained diffusion model to propagate keypoints across an entire dataset, significantly reducing the cost of manual annotation. Additionally, we enhance the Vision Transformer (ViT) by implementing Keypoint Positional Encoding (KPE) and Categorical Keypoint Positional Embedding (CKPE), enabling the ViT to learn more robust and semantically-aware representations. This provides more comprehensive and detailed keypoint representations, leading to more accurate and efficient re-identification. Our extensive experimental evaluations demonstrate that this approach significantly outperforms existing state-of-the-art methods across four wildlife datasets. The code will be publicly released.

\end{abstract}    
\section{Introduction}
\label{sec:intro}

% Animal re-identification (ReID) has emerged as a critical tool in ecological research, substantially contributing to the monitoring of wildlife populations, behavioral analysis, and ecological impact assessments, which are essential for effective conservation strategies \cite{schneider2022similarity,schofield2022more,vidal2021perspectives}. Unlike pedestrian ReID, animal ReID in natural settings is fraught with challenges due to limited availability of annotated data, diverse environmental conditions, and frequent occlusions that obscure crucial visual information. Compounded by intrinsic variations in animal shapes, postures, and patterns, these factors significantly complicate the ReID process \cite{ravoor2020deep,li2019atrw}.

Animal re-identification (ReID) has emerged as a critical tool in ecological research, providing indispensable insights into the monitoring of wildlife populations, behavioral analysis, and ecological impact assessments. These insights are foundational to developing effective conservation strategies \cite{schneider2022similarity,schofield2022more,vidal2021perspectives}. Unlike pedestrian ReID, animal ReID in natural settings presents formidable challenges that stem from the limited availability of annotated data, highly variable environmental conditions, and inability to directly apply pre-trained models to animal data~\cite{kreiss2019pifpaf} that obscure crucial visual information. These challenges are compounded by intrinsic variations in animal shapes, postures, and patterns, which significantly complicate the ReID process \cite{ravoor2020deep,li2019atrw}.

\begin{figure}[t!]
    \centering
    \includegraphics[width=0.9\linewidth]{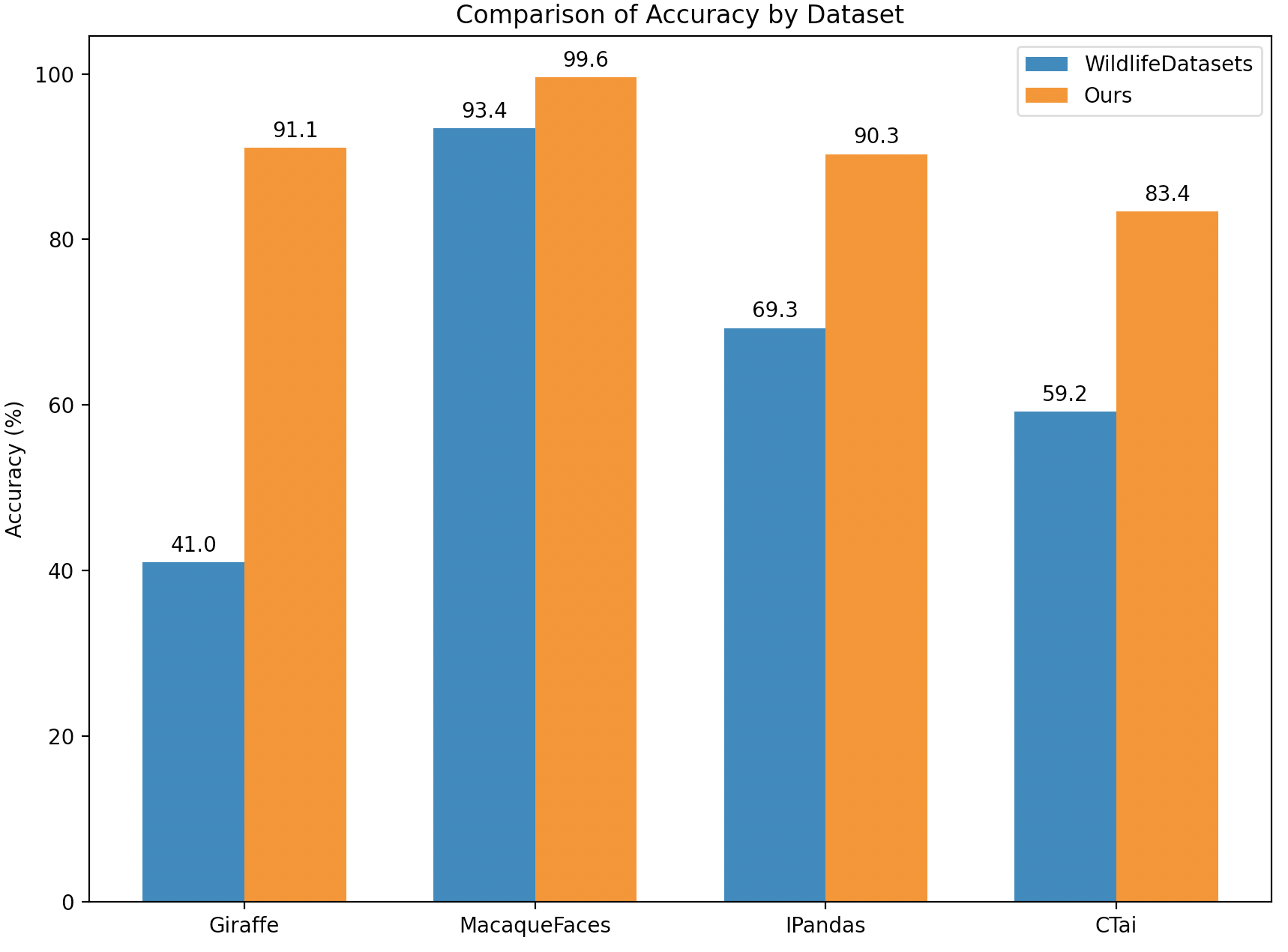}
    \caption{Comparison of model performance between the previous state-of-the-art (SOTA) and our proposed method across four datasets. Our approach significantly outperforms the previous SOTA, demonstrating notable improvements in accuracy for all datasets.}
    \label{fig:first}
\end{figure}

% Recent advancements in animal ReID methodologies, although varied, share the unified goal of accurately and efficiently recognizing individual animals within a species based on unique phenotypic traits such as markings, patterns, or other distinctive features \cite{li2019atrw,crall2013hotspotter}. Conventional methods often require manual selection of features at specific anatomical locations—such as ears or fins—which is both labor-intensive and necessitates expert knowledge, rendering it impractical for large wildlife datasets \cite{lahiri2011biometric}. Some study underscored the importance of keypoints in ReID tasks through the introduction of the Keypoint-Promptable Transformer (KPR) for human ReID, demonstrating the pivotal role of keypoints \cite{somers2025keypoint}. However, these methods are primarily effective on human-centric datasets and do not translate well to the anatomical and behavioral diversity of animals. The advent of diffusion feature has enabled effective keypoint propagation across images within a dataset, marking a significant breakthrough in the field \cite{tang2023emergent}.

\begin{figure}[th]
    \centering
    \includegraphics[width=1.0\linewidth]{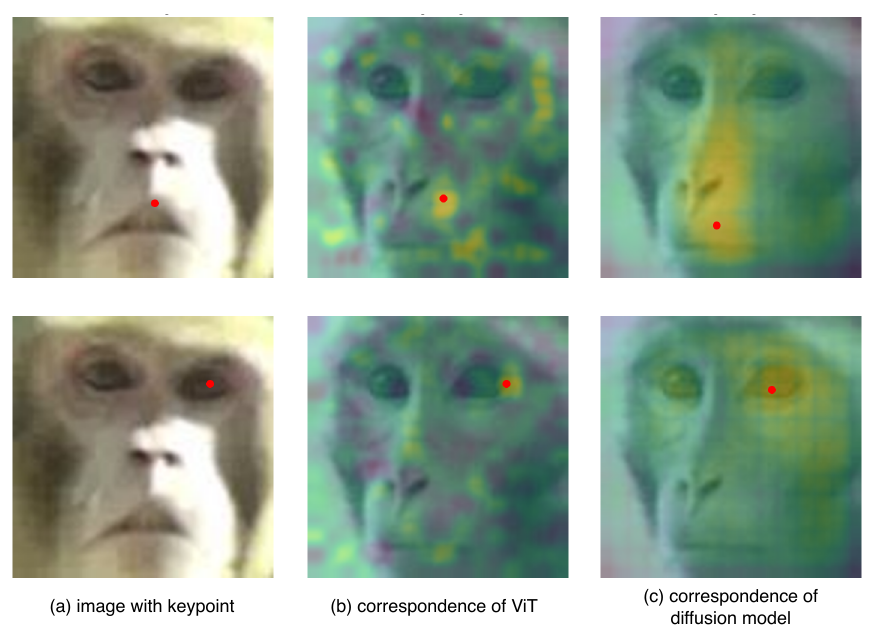}
    \caption{Comparison of keypoint similarity heatmaps between a pre-trained Vision Transformer (ViT) model and a pre-trained diffusion model. The most salient feature in the heatmap, indicating the keypoint, is highlighted. (a) is the image and the keypoint, above is the mouth and bottom is the right eye. (b) is the similarity heatmaps from ViT and (c) is the similarity heatmaps from Stable Diffusion. It is obvious that the pre-trained diffusion model exhibits stronger semantic correspondence compared to the pre-trained ViT model.}
    \label{fig:vit_vs_dif}
\end{figure}

Advancements in animal ReID methodologies, though diverse, share a unified goal of accurately and efficiently recognizing individual animals within a species based on unique phenotypic traits such as markings, patterns, or other distinctive features \cite{li2019atrw,crall2013hotspotter}. Traditional approaches frequently require manual selection of features at specific anatomical locations—such as ears or fins—which is labor-intensive and requires extensive expert knowledge, rendering these methods impractical for processing large wildlife datasets \cite{lahiri2011biometric}. Recent studies have underscored the importance of keypoints in ReID tasks, as demonstrated through the introduction of the Keypoint-Promptable Transformer (KPR) for human ReID, highlighting the pivotal role of keypoints in enhancing identification accuracy \cite{somers2025keypoint}. However, these methods are primarily effective on human-centric datasets and face significant challenges when applied to the anatomical and behavioral diversity of animals. The emergency of corresponding diffusion features has significantly enhanced the capability for effective keypoint propagation across images within a dataset, thereby facilitating this process.\cite{tang2023emergent}, e.g., as demonstrated in Figure~\ref{fig:vit_vs_dif}.

Motivated by these insights, our study introduces Keypoint Propagation, a novel mechanism where keypoints annotated on a single image by GPT-4 can be automatically propagated throughout the entire dataset, utilizing a pre-trained diffusion model that adapts to the significant variability inherent in wildlife imagery. We further propose Keypoint Positional Embedding (KPE) and Categorical Keypoint Positional Embedding (CKPE). These two novel Embedding strategies enhance the capabilities of the Vision Transformer (ViT), enabling it to recognize and interpret the semantic relevance of keypoints. In specific, The KPE method encodes the spatial information of keypoints into a high-dimensional space, aiding the ViT in focusing on these positions during the learning process. Extending this, our CKPE approach incorporates categorical data about the keypoints, providing the model with semantic awareness necessary for distinguishing closely related species or individuals within species. Our primary contributions are as follows:

\begin{itemize}

    \item We introduce a diffusion-based keypoint propagation mechanism that minimizes the need for extensive annotations.

    \item We propose Keypoint Positional Embedding (KPE) and Categorical Keypoint Positional Embedding (CKPE) to enhance the Vision Transformer's (ViT) ability to recognize and contextualize keypoints effectively.
    
    \item Our methods outperform existing state-of-the-art approaches in occluded ReID tasks, as evidenced by significant accuracy improvements ranging from +5.9\% to +50.1\% on four publicly available benchmarks (shown in Fig~\ref{fig:first}), respectively.

\end{itemize}

To facilitate further research and practical applications in this field, we will made our codebase, datasets, and detailed annotations publicly available, providing a comprehensive framework for the evaluation of promptable wildlife ReID methods.

\section{Related Work}
\label{sec:related_work}

% \subsection{General ReID Models}
% Re-Identification (ReID) technologies, traditionally focused on humans and vehicles, utilize a range of approaches from hand-crafted features to deep learning architectures. Early efforts in the field demonstrated that metric learning and deep feature extraction are effective for managing the variability and complexity of ReID tasks \cite{zheng2016person, liu2017survey}. Innovations such as the integration of siamese networks and attention mechanisms have significantly advanced the field \cite{varior2016gated, li2018harmonious}.

\subsection{Animal ReID}
Recent advances in deep learning, particularly in convolutional neural networks (CNNs), have significantly advanced wildlife re-identification (ReID), with notable contributions in feature extraction \cite{ahmed2015improved,li2014deepreid} and metric learning \cite{liao2015person,xiong2014person}. While person ReID has reached a certain level of maturity, wildlife ReID is still in its nascent stages, with many existing methods being species-specific, thereby limiting their broader applicability \cite{norouzzadeh2018automatically, matthe2017comparison,halloran2015applying}. Current approaches can be broadly categorized into three main strategies: (1) \textit{Global Feature Learning}, which utilizes whole animal images to extract global features. These methods, such as the multi-stream feature fusion network for giant pandas \cite{wang2021giant} and the specialized loss for manta ray ReID \cite{moskvyak2021robust}, are adaptable across species but often rely on techniques originally developed for person ReID. (2) \textit{Species-specific Feature Extraction}, which targets distinct anatomical features, such as dolphin fins \cite{bouma2018individual,konovalov2018individual,weideman2017integral}, whale tails \cite{cheeseman2022advanced}, and elephant ears \cite{weideman2020extracting}. These methods, while effective within a specific species, are limited by challenges such as occlusions and viewpoint variations, which hinder their generalizability across species. (3) \textit{Auxiliary Information Integration}, which incorporates pose keypoint estimation to enhance feature learning, as demonstrated in tiger \cite{li2019atrw} and yak \cite{zhang2021yakreid} ReID. While these approaches can improve accuracy, they rely on auxiliary data, making them difficult to standardize across diverse species. Recently, the WildlifeDatasets toolkit \cite{Cermak_2024_WACV} was introduced to establish a standard baseline for animal ReID, alongside several general approaches\cite{jiao2024toward,shinoda2025petface} and datasets aimed at developing a universal and robust solution for wildlife ReID. Despite these advancements, there remains a critical need for a clearly defined task and high-performance solutions that can generalize across a wide range of species.

% \subsection{Diffusion and Keypoint Embedding in ReID}
% Diffusion techniques have been explored in other areas of computer vision for their efficacy in propagating information across a dataset, a feature that is invaluable in ReID for extending limited annotations \cite{he2020momentum, chen2020simple}. The integration of these techniques with keypoint embedding as discussed by Tian et al. \cite{tian2020makes} further facilitates the identification process by enhancing the semantic correspondence across images. Our approach builds on these methodologies by employing diffusion-based feature matching and embedding within a Vision Transformer architecture to improve the scalability and accuracy of animal ReID.

\subsection{ViT for ReID}

While convolutional neural network (CNN)-based methods have long dominated person and vehicle re-identification (ReID), the introduction of the Vision Transformer (ViT) has significantly advanced the field \cite{ye2024transformer}. In general object ReID, models such as TransReID \cite{he2021transreid}, which is entirely ViT-based, have shown improved cross-camera ReID accuracy by emphasizing local feature learning and viewpoint information. Additionally, several hybrid models \cite{li2022pyramidal,li2021diverse,zhang2021hat} combine ViT with CNNs, with techniques like dual cross-attention learning \cite{zhu2022dual} enhancing both global and local feature learning through more refined attention mechanisms. Some methods \cite{wang2022pose} also leverage auxiliary information, such as pose estimation, to improve the extraction of human body-related features. \cite{somers2025keypoint} introduced keypoints from pre-trained human keypoint detection models, demonstrating the importance of auxiliary prompts. However, these prompts are only effective on human datasets and do not generalize well to wildlife images due to the increased complexity of animal scenes. In this paper, we propose a novel mechanism for detecting and utilizing keypoints specifically in wildlife images.

\subsection{Visual foundation model with correspondences}
 Recent advancements in visual foundation models, such as Vision Transformers (ViTs) and Stable Diffusion (SD), have demonstrated that pretrained features—learned through classification tasks or generative tasks \cite{amir2021deep, amir2022effectiveness, parmar2023zero, he2020momentum}—serve as powerful descriptors for semantic matching, often outperforming previous methods specifically designed for this task. Stable Diffusion, trained on billions of text-image pairs \cite{schuhmann2022laion}, has led to the development of widely adopted open-source text-to-image models. Studies \cite{tang2023emergent, hedlin2024unsupervised} have shown that correspondences can emerge in image diffusion models without explicit supervision, thereby suggesting the potential for the keypoint propagation within these models.

\begin{figure*}[th!]
    \centering
    \includegraphics[width=1.0\linewidth]{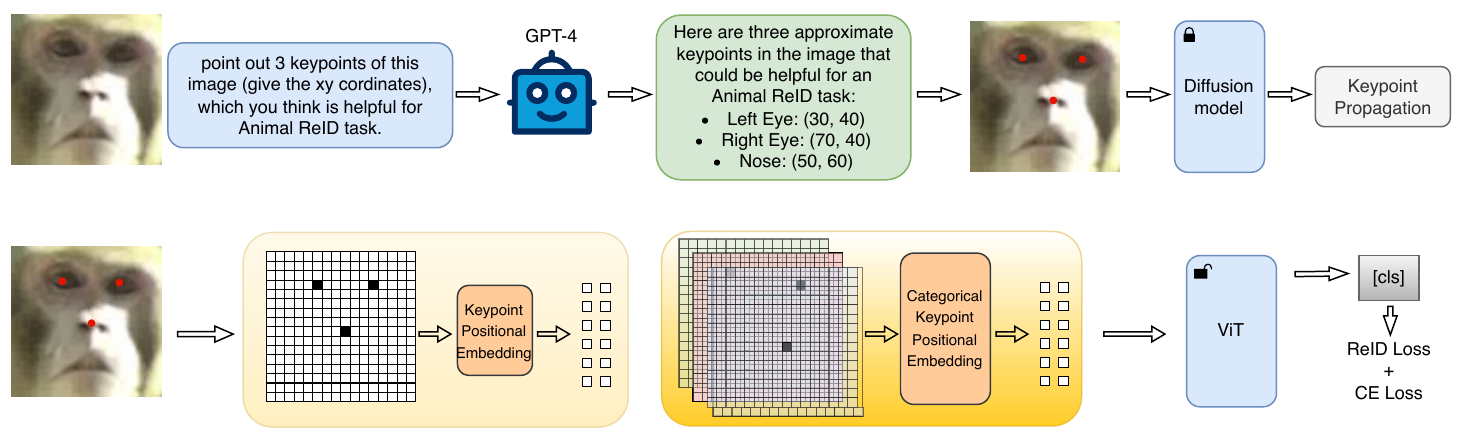}
    \caption{Architecture Overview of Our Proposed Categorical Keypoint Positional Embedding for Wild Animal ReID: Initially, GPT-4 identifies critical keypoints for the ReID task from a single image. Subsequently, a diffusion model is employed to propagate these keypoints across the entire dataset. The identified keypoints then inform our ViT-based ReID module through Keypoint Positional Embedding or Categorical Keypoint Positional Embedding, focusing on the content at these specific locations and semantic information to enhance feature discrimination and accuracy. }
    \label{fig:main}
\end{figure*}

\section{Method}
\label{sec:method}

In this section, we start from task definition (Sec.~\ref{subsec:kp_0}) and then introduce Keypoint Detection and Propagation (Sec.~\ref{subsec:kp_1}), Keypoint Positional Embedding (Sec.~\ref{subsec:kp_2}), Categorical Keypoint Positional Embedding (Sec.~\ref{subsec:kp_3}), and entire training testing process (Sec.~\ref{subsec:kp_4}). The comprehensive framework of our method is depicted in Figure~\ref{fig:main}.

\subsection{Task forming}
\label{subsec:kp_0}

Animal Re-Identification (ReID) is a computer vision task aimed at identifying and matching individual animals across different non-overlapping camera captures. We formally define this task as follows:

\textbf{Given:}
\begin{itemize}
    \item A \textbf{training set} \(\mathcal{T}\) used to train the model, which includes images and their ID labels.
    \item A \textbf{query set} \(\mathcal{Q}_{\text{test}} = \{q_1, q_2, \dots, q_n\}\) and a \textbf{gallery set} \(\mathcal{G}_{\text{test}} = \{g_1, g_2, \dots, g_m\}\) for testing. The training and testing sets are distinct, ensuring no overlap to simulate real-world scenarios where the model encounters completely unseen samples. This separation tests the model's generalization capabilities and effectiveness in practical applications.
\end{itemize}

\textbf{Objective:}
\begin{itemize}
    \item Learn a mapping function \(f: \mathcal{I} \rightarrow \mathbb{R}^d\) that maps an image \(I\) to a point in a \(d\)-dimensional Euclidean space. This function is optimized so that images of the same animal are closer together, and those of different animals are farther apart.
    \item For each query image \(q_i \in \mathcal{Q}_{\text{test}}\), find the set \( \{g_j \in \mathcal{G}_{\text{test}} : \text{identity}(q_i) = \text{identity}(g_j)\} \), using the trained model to predict and match identities based on learned representations.
\end{itemize}

\subsection{Keypoint Detection and Propagration}
\label{subsec:kp_1}
Given an input image, we begin by leveraging GPT-4 to identify keypoints with high discriminative power, which are critical for the re-identification (ReID) task. These keypoints, identified by GPT-4, capture critical spatial locations that distinguish different instances of the same species. Once identified, these keypoints are mapped to corresponding locations across the entire dataset using a pre-trained stable diffusion model. Specifically, let \( p_i \) represent the location of the \(i\)-th keypoint in an image \(I\), where \( i \in [1, N] \) and \( N \) denotes the total number of keypoints. Using the diffusion model, matching keypoints across all images in the dataset are retrieved, creating a consistent set of keypoints for each image, denoted as \( \{ p_1, p_2, \dots, p_N \} \).

The process of keypoint detection and propagation is crucial in mitigating the data sparsity issue commonly faced in wildlife ReID, where manual annotation is expensive and often impractical. To compute the feature similarity map between two images \( I^{ref} \) (the reference image) and \( I^{target} \) (the target image), we first extract their respective diffusion features, \( \mathbf{F}^{ref} \) and \( \mathbf{F}^{target} \). For each keypoint location \( p_i \), defined by the coordinates \( (x_i, y_i) \), we extract the corresponding feature vector \( v_{p_{i}} = F^{ref}_{x_i, y_i} \) from the reference image. These feature vectors are normalized to unit length, resulting in normalized feature vectors \( \hat{\mathbf{v}}_{p_{i}} \).

To compute the similarity map between keypoints and target image, we define the cosine similarity map between the normalized feature vector at the keypoint \( p_n \) in the reference image and the feature map of the target image. The cosine similarity is expressed as:

\[
S_{n} = \hat{\mathbf{v}}_{p_{n}} \cdot \mathbf{F}^{target}
\]

\noindent where \( S_{n} \) represents the similarity map between the feature vector at \( p_n \) and the target image. This computation yields a similarity map \( \mathbf{S} \) for each keypoint location in the reference image, where the point with the highest similarity score in \( S_n \) corresponds to the matching keypoint location in the target image.

\[
p_{n} = argmax\  S_{n}
\]

By this, we can propagate the keypoints across the dataset effectively, ensuring consistent keypoint localization even when faced with varying poses, occlusions, and environmental conditions. By utilizing both GPT-4 for initial keypoint detection and stable diffusion models for propagation, we significantly reduce the need for manual annotations while enhancing the ReID model's ability to handle the variability present in real-world wildlife imagery.

\subsection{Keypoint Positional Embedding (KPE)}
\label{subsec:kp_2}

We introduce the \textbf{Keypoint Positional Embedding (KPE)} algorithm, designed to enhance the Vision Transformer (ViT) by integrating both visual and spatial information derived from keypoints in an image. Given an image \( I \) with a set of keypoints \( \{ p_1, p_2, \dots, p_N \} \), the KPE algorithm works by mapping the identified keypoints \( p \) into a one-hot vector mask representation. This mask representation is designed to encode the spatial presence of each keypoint relative to the patches of the image as used by the ViT model. Specifically, for image \( I \), we create the mask vector \( \mathbf{M}_I \), where each element of the vector corresponds to a patch in the image. We define the mask as follows:

\[
\mathbf{M}_{I} = \text{OneHot}(p) \in \mathbb{R}^{K+1}
\]

\noindent where \( K \) is the total number of image patches (corresponding to the patch configuration of ViT), and \( +1 \) accounts for the inclusion of the [CLS] token in the ViT architecture. For each patch in the image, if it contains the keypoint \( p \), the corresponding value in the mask is set to 1; otherwise, it is set to 0. This one-hot encoding efficiently captures the spatial distribution of the keypoints across the image and ensures that the ViT can learn from both the content of the patches and the locations of the keypoints.

In the KPE approach, we introduce a learnable weight matrix \( W_{\text{kp}} \in \mathbb{R}^{(K+1) \times d} \), where \( d \) is the embedding dimension. The matrix \( W_{\text{kp}} \) serves to project the one-hot vector mask into a high-dimensional embedding space, allowing the model to capture the spatial relationships between the keypoints and the image patches. For each keypoint mask \( \mathbf{M}_i \), the corresponding embedding vector is computed by the following matrix multiplication:

\[
\mathbf{M}_i \cdot W_{\text{kp}} \in \mathbb{R}^d
\]

This embedding vector is then added to the image patch features extracted by the ViT model. By incorporating the spatial information of the keypoints into the ViT’s attention mechanism, the KPE algorithm enables the model to focus on regions of the image that are most relevant for the ReID task, while also leveraging the semantic understanding provided by the keypoints. The integration of this keypoint embedding allows the ViT to learn both the visual patterns and the spatial positioning of keypoints, improving the model's ability to handle complex real-world scenarios where precise localization and recognition are crucial.

\subsection{Categorical Keypoint Positional Embedding (CKPE)}
\label{subsec:kp_3}

Building upon the Keypoint Positional Embedding (KPE), we propose the \textbf{Categorical Keypoint Positional Embedding (CKPE)}, which enhances the model by incorporating category-specific information for each keypoint. The inclusion of this categorical information is particularly useful for tasks where different keypoints may represent distinct anatomical or semantic features, such as facial landmarks or body parts. For instance, in the case of a monkey face dataset, let \( C_i \) denote the category of the \( i \)-th keypoint \( p_i \), where \( C_i \in \{\text{left eye}, \text{right eye}, \text{nose}\} \). This categorization allows the model to differentiate between various types of keypoints within the same image, providing it with more granular semantic information.

To encode this categorical information, we construct a categorical mask for the keypoints in the image. The categorical mask \( \mathbf{CM}_I \) is a one-hot vector representation that maps each keypoint \( p_i \) to its corresponding category. Specifically, we define the categorical mask as follows:

\[
\mathbf{CM}_{I} = \text{OneHot}(P_I) \in \mathbb{R}^{(K+1) \times N_c}
\]

\noindent where \( K \) is the total number of image patches, which corresponds to the patch configuration in the ViT model, and \( N_c \) is the number of categories associated with the keypoints (e.g., for the monkey face dataset, \( N_c = 3 \), corresponding to "left eye", "right eye", and "nose"). For each patch in the image, if it contains the keypoint \( p_i \) of category \( C_i \), the corresponding value in the mask is set to 1; otherwise, it is set to 0. This one-hot encoding allows the model to learn not only the spatial distribution of keypoints but also their specific categories, enriching the contextual understanding of each keypoint's role in the image.

To further enhance the model’s ability to learn category-specific features, we introduce a learnable category-specific embedding matrix \( W_{\text{ckp}} \in \mathbb{R}^{N_c \times d} \), where \( N_c \) is the number of categories and \( d \) is the embedding dimension. This matrix projects the one-hot vector representation of the categorical mask into a higher-dimensional space, allowing the model to capture the relationships between keypoints and their respective categories. The categorical embedding for keypoints in image \( I \) is then computed as:

\[
\mathbf{CM}_I \cdot W_{\text{ckp}} \in \mathbb{R}^d
\]

This embedding vector, representing the categorical information of the keypoints, is subsequently added to the image patch features extracted by the ViT model. The integration of categorical embeddings into the ViT’s attention mechanism allows the model to not only focus on the spatial location of keypoints but also learn their semantic significance, such as distinguishing between different facial features in the case of monkey faces or differentiating body parts in animal datasets.

The advantage of CKPE lies in its ability to provide both spatial and semantic context for each keypoint, which is particularly beneficial for tasks such as wildlife ReID or human pose estimation, where the model must differentiate between various types of keypoints (e.g., eyes, ears, nose) that belong to the same individual or species. By embedding categorical data alongside positional information, CKPE enables the ViT to learn more robust and semantic-aware representations, providing a more comprehensive and nuanced representation of keypoints, ultimately leading to more accurate and efficient re-identification.

% In summary, CKPE not only enhances the model's capacity to capture the spatial positioning of keypoints but also injects crucial semantic understanding, allowing for improved ReID performance in both wildlife and human identification tasks. By incorporating category-specific information into the ViT’s learning process, CKPE provides a more comprehensive and nuanced representation of keypoints, ultimately leading to more accurate and efficient re-identification.

\subsection{Training and Testing}

\label{subsec:kp_4}

Once the keypoint and categorical embeddings have been integrated into the Vision Transformer (ViT), we proceed with fine-tuning the model for the re-identification task. Specifically, we use the [CLS] token as a global feature representation, allowing the model to aggregate visual, positional and semantic information from the keypoints, including their spatial and categorical relationships. This enables the ViT to simultaneously learn discriminative features from the keypoints and their positional contexts within the image. The objective is for the model to learn a rich representation that not only captures the individual characteristics of each keypoint but also the semantic importance of its category, thereby improving the overall performance of the re-identification task.

For training, we define a composite loss function that consists of two primary components: the classification loss and the re-identification loss. The total loss function is given by:

\[
\mathcal{L} = \mathcal{L}_{\text{reid}} + \lambda \mathcal{L}_{\text{ce}}
\]

\noindent where the re-identification loss, \( \mathcal{L}_{\text{reid}} \), is ArcFace loss based on a triplet loss function for re-identification task. \( \mathcal{L}_{\text{ce}} \) represents the classification loss, which is cross entropy loss, weighted by a scalar \( \lambda \) to balance the two objectives.

During the training phase, the model simultaneously learns to classify keypoints correctly, predict their categories, and distinguish between different identities in the dataset. The [CLS] token serves as the aggregated feature representation for each image, which encapsulates both the visual and semantic content, including the keypoint features and their respective categories. This joint optimization ensures that the learned representation effectively supports both the spatial and categorical aspects of the task.

For testing, we use the [CLS] embedding extracted from the trained ViT model. During the testing phase, the model evaluates the similarity between the query image and the gallery of test images, which consist of individuals never seen during the training process. To perform the matching, we compute the cosine similarity between the [CLS] embedding of the query image and the embeddings of the gallery images. Specifically, for a given query image \( I^q \) and a set of gallery images \( \{ I^g_1, I^g_2, \dots, I^g_N \} \), we compute the cosine similarity \( S \) as:

\[
S = \frac{\mathbf{z}^q \cdot \mathbf{z}^{\mathcal{G}}} {\| \mathbf{z}^{q} \|_2 \| \mathbf{z}^{\mathcal{G}} \|_2}
\]

\noindent where \( \mathbf{z}^q \) is the [CLS] embedding of the query image and \( \mathbf{z}^\mathcal{G} \) is the [CLS] embeddings of gallery images. The gallery image with the highest similarity score to the query image is selected as the most likely match.

\section{Experiment}
\label{sec:expe}

In this section, we first provide details of our experimental setup. Next, we assess our proposed method on four widely used Animal ReID benchmarks, comparing it against previous SOTA methods. Lastly, we conduct comprehensive ablation studies to further explore the effectiveness of our approach.

\begin{table*}[h!]
\centering
\begin{tabular}{c|cccc}
\hline
                 & MacaqueFaces & Giraffe & IPandas &CTai \\ \hline
WildLife Dataset\cite{Cermak_2024_WACV} & 93.4\%       & 41.0\%  & 69.3\% & 59.2\%  \\
ours (KPE)       & 99.1\%       & 89.6\%  & 88.2\%  & 81.6\%  \\
ours (CKPE)      & 99.6\%       & 91.1\%  & 90.3\%  & 83.4\% \\ \hline
\end{tabular}
\caption{Quantitative comparisons between our methods and previous SOTA on the 4 public wildlife ReID benchmarks are presented. Our approach surpasses wildlife dataset and shows effectiveness of our KPE and CKPE.}
\label{tab:main}
\end{table*}

\subsection{Experimental Setup}
\textbf{Dataset:} 
In our study, we assess our method using four popular public datasets for Animal ReID tasks: MacaqueFaces~\cite{witham2018automated} and Giraffe~\cite{miele2021revisiting}. “MacaqueFaces" dataset comprises 6,280 photographs of 81 individual rhesus macaques, designing for facial recognition and animal ReID. Similarly, the "Giraffe" dataset includes 1,393 images of 178 individual giraffes. The CTai~\cite{freytag2016chimpanzee} dataset contains 4,662 images of chimpanzee faces from 71 individuals, while the IPandas~\cite{wang2021giant} dataset includes 6,874 images of 50 pandas. All the data split is 80\% of individuals for training, and 20\% for testing.

\noindent \textbf{Evaluation Metrics:} Follow \cite{Cermak_2024_WACV, shinoda2024petface}, we report the accuracy. In particular, Accuracy = $ \frac{TP + TN}{TP + TN + FP + FN} $,  where 
TP (True Positives) is the count of correctly identified positive cases. TN (True Negatives) is the count of correctly identified negative cases. FP (False Positives) is the count of negative cases incorrectly identified as positive. FN (False Negatives) is the count of positive cases incorrectly identified as negative.

All experiments in this work are conducted on one NVIDIA GeForce RTX 3090 GPU and all other experimental settings (except the data division) are identical to the WildlifeDatasets ~\cite{Cermak_2024_WACV} approach for a fair comparison. $\lambda$ in $\mathcal{L}$ is set to 0.2 empirically.

\subsection{Comparison with State-of-the-Arts}

Table~\ref{tab:main} presents a comprehensive comparison of our method against the state-of-the-art (SOTA) on four popular Wildlife ReID benchmarks. Our approach demonstrates consistent and substantial performance improvements across all datasets. Specifically, on the Giraffe dataset, our method achieves an accuracy of 91.1\%, representing a significant improvement over the previous SOTA, which achieved only 41.0\%. On the other three publicly available datasets—MacaqueFaces, IPandas, and CTai—our method outperforms the previous SOTA by 6.2\%, 21.0\%, and 24.2\%, respectively. These results highlight the robustness and generalizability of our approach across different species and environments. Furthermore, the performance gain observed when using our \textbf{Categorical Keypoint Positional Embedding (CKPE)} over \textbf{Keypoint Positional Embedding (KPE)} underscores the effectiveness of incorporating both spatial information and semantic categories of keypoints, which plays a crucial role in improving model accuracy and robustness in wildlife ReID tasks.

\section{Ablation study}
\label{sec:ablation-study}

In this section, we are interested in ablating our approach from the following perspective views on MacaqueFaces dataset (default size of ViT is large):

\subsection{Different approaches to find keypoints}
% We compared the following approaches to find keypoints for the single images: "Grad-CAM (ResNet)", "Grad-CAM (ViT)", "Experts (from School of Biological Sciences)" and "GPT-4".

\begin{table}[h!]
\centering
\resizebox{0.95\columnwidth}{!}{
\begin{tabular}{c|cccc}
\hline
Methods    & \begin{tabular}[c]{@{}l@{}}Grad-CAM\\ (ResNet)\end{tabular} & \begin{tabular}[c]{@{}l@{}}Grad-CAM\\ (ViT)\end{tabular} & Expert & GPT-4   \\ \hline
kp  &\includegraphics[width=1.7cm, height=1.7cm]{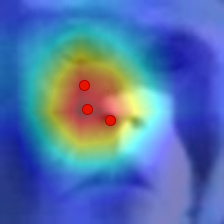}&\includegraphics[width=1.7cm, height=1.7cm]{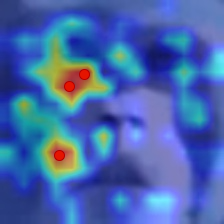}&\includegraphics[width=1.7cm, height=1.7cm]{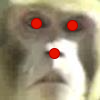}&\includegraphics[width=1.7cm, height=1.7cm]{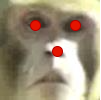} \\ \hline
Acc & 96.9\%   & 97.7\%   & 99.20\% & 99.20\% \\ \hline
\end{tabular}
}
\caption{Effectiveness of different approaches of defining 3 keypoints for single image.}

\label{tab:kpdef} 
\end{table}

% Table~\ref{tab:kpdef} shows the keypoints positions and their accuracy from different methods, in specific, the CAM of ResNet focus on the nose area, the CAM of ViT seems to random points, while experts and GPT-4 focus on distinctive area (eyes, nose). Accuracy also proves the effectiveness of GPT-4's position, which has the same performance with expert's keypoint and surpasses the CAM from both ResNet and ViT.

In this ablation study, we systematically compare several keypoint detection approaches to evaluate their performance in identifying discriminative regions in single images. We investigate four different methods: \textbf{Grad-CAM (ResNet)}, \textbf{Grad-CAM (ViT)}, \textbf{Experts (from the School of Biological Sciences)}, and \textbf{GPT-4}. Each method brings unique characteristics and strengths, which we examine in the context of wildlife re-identification. The goal is to assess the accuracy and consistency of keypoint localization across various detection algorithms and identify the most effective approach for this task.

\begin{itemize}
    \item \textbf{Grad-CAM (ResNet)}: Grad-CAM (Gradient-weighted Class Activation Mapping) applied to the ResNet architecture generates heatmaps by using the gradients of the target class with respect to the final convolutional layer. Grad-CAM highlights areas of the image that contribute most to the classification decision. In our case, Grad-CAM applied to ResNet tends to focus on the general area around the nose, capturing some key features of the face but with limited precision. 
    
    \item \textbf{Grad-CAM (ViT)}: In contrast to ResNet, Grad-CAM applied to the Vision Transformer (ViT) visualizes attention maps from transformer-based models. ViT leverages self-attention mechanisms that enable the model to focus on different parts of the image. However, when applied to keypoint detection, Grad-CAM on ViT appears to focus on scattered and random points across the image, making it less effective for precise keypoint localization. While ViT excels in capturing global relationships within an image, its application to keypoint localization in wildlife images shows some limitations in pinpointing accurate anatomical features.
    
    \item \textbf{Experts (from the School of Biological Sciences)}: To establish a benchmark for keypoint detection, we incorporate expert annotations from professionals in biological sciences. These experts, with extensive domain-specific knowledge, manually identify keypoints based on anatomical landmarks. For example, in the case of monkey face images, experts can pinpoint key features such as the eyes, nose, and mouth with high accuracy. This provides a reliable and accurate ground truth for keypoints comparison.
    
    \item \textbf{GPT-4}: Lastly, we explore the use of GPT-4, a state-of-the-art language model that has been shown to perform well in multi-modal tasks involving both vision and language. By leveraging GPT-4’s advanced understanding of contextual relationships, we apply it to identify keypoints in wildlife images. GPT-4 shows remarkable accuracy in detecting keypoints' locations, focusing on the most distinctive areas of the image, such as the eyes and nose. In this study, GPT-4 performs on par with expert annotations, demonstrating its ability to effectively localize keypoints without requiring explicit image labeling or manual intervention.
\end{itemize}

Table~\ref{tab:kpdef} summarizes the keypoint positions and the corresponding accuracy achieved by each method. The heatmaps generated by Grad-CAM on ResNet predominantly highlight the nose region, though it lacks the precision required for reliable keypoint localization. On the other hand, Grad-CAM on ViT appears to focus on random and scattered regions of the face, resulting in a less focused and less accurate keypoint detection. In contrast, both expert annotations and GPT-4 successfully localize keypoints in more distinctive regions such as the eyes and nose, providing more precise and relevant keypoint positions. 

The accuracy results further support the effectiveness of GPT-4 in keypoint detection. Specifically, the performance of GPT-4 matches that of the expert annotations, demonstrating its high accuracy in identifying keypoints. Furthermore, GPT-4 outperforms both ResNet and ViT-based Grad-CAM methods, surpassing their accuracy by a significant margin. These findings highlight the potential of GPT-4 as a powerful tool for keypoint localization, especially in tasks where manual annotations are unavailable or impractical.

% \subsection{Different timesteps to propagate the keypoints}
% In Stable Diffusion, different time step has different feature, we compared the performance with keypoints from different timesteps.

\begin{table*}[ht]
\centering

\resizebox{1.9\columnwidth}{!}{
\begin{tabular}{l|lllllllll}
\hline
Time step & 0      & 50     & 100    & 150    & 200    & 250   &300 & 500    & 1000   \\ \hline
Acc       & 98.9\% & 98.9\% & 98.2\% & 98.9\% & 99.2\% & 99.2\% & 99.2\% & 97.2\% & 96.9\% \\ \hline
\end{tabular}
}
\caption{Comparison between performance of keypoint propagation from different timesteps diffusion feature.}
\label{tab:timestep} 
\end{table*}

\subsection{Different Timesteps to Propagate the Keypoints}

In Stable Diffusion, the process of keypoint propagation is influenced by the timestep at which the diffusion model operates. Each timestep in the diffusion process represents a stage in the image generation or feature transformation, with earlier timesteps capturing more coarse, global features and later timesteps refining those features into more detailed representations. To investigate the effect of different timesteps on keypoint propagation, we compare the performance of keypoints extracted from multiple timesteps during the diffusion process.

Table~\ref{tab:timestep} presents the performance of keypoint propagation at different diffusion timesteps. Specifically, we evaluate the accuracy of keypoint localization when propagated from the 0-300 timestep range, compared to the performance at later stages such as timestep 500 and timestep 1000. The results indicate that keypoints propagated from timesteps in the range of 0-300 yield stable and competitive performance, aligning with similar findings reported in \cite{}. These early timesteps appear to capture the most significant features related to keypoint locations, offering a balanced trade-off between feature accuracy and computational efficiency.

On the other hand, timesteps 500 and 1000, which correspond to later stages in the diffusion process, show diminishing returns in terms of performance when compared to the 0-300 timestep range. This suggests that while later timesteps may provide increasingly detailed features, they do not significantly improve the localization of keypoints in this particular task, and may even introduce noise or irrelevant details that detract from the model's focus on the keypoint positions.

% Table~\ref{tab:timestep} shows the performance of keypoints propagation from different timesteps diffusion feature. Similar to the findings in \cite{}, time step 0-300 obtain stable and competitive performance compared to 500 and 1000.

% \subsection{Different ViT Scales}
\begin{table}[ht]
\centering
\resizebox{0.9\columnwidth}{!}{

\begin{tabular}{cccc}

\hline
ViT Scale      & W/O keypoints   & KPE    & CKPE   \\ \hline
      
base  & 92.5\% & 98.5\% & 97.3\% \\
large & 96.9\% & 98.9\% & 99.2\% \\
huge  & 98.9\% & 99.1\% & 99.6\% \\ \hline
\end{tabular}
}
\caption{Effectiveness of our KPE and CKPE on different scales of ViTs.}
\label{tab:vitsize} 

\end{table}

% Table~\ref{tab:vitsize} shows the effectiveness of KPE and CKPE on different scales of ViTs from base to huge. It is clear that our KPE and CKPE both have solid improvement on all scales of ViTs.
\subsection{Different ViT Scales}

 We also evaluate the effectiveness of Keypoint Positional Embedding (KPE) and Categorical Keypoint Positional Embedding (CKPE) across different scales of Vision Transformers (ViTs), ranging from ViT-Base to ViT-Huge. As shown in Table~\ref{tab:vitsize}, both KPE and CKPE consistently improve performance across all ViT scales. The improvements are particularly noticeable in the smaller ViT-Base models, where the embeddings help overcome the model's limited capacity to model complex relationships. As the scale increases to ViT-Large and ViT-Huge, the performance gains remain significant, indicating that KPE and CKPE continue to enhance the model's ability to refine spatial and categorical keypoint information, benefiting from the increased capacity of the larger models. These results demonstrate that KPE and CKPE offer scalable performance improvements, complementing the strengths of ViTs regardless of model size, making them effective across a range of ViT configurations, from base to huge.

\subsection{Different numbers of keypoints}

\begin{table}[ht]
\centering
\resizebox{1.0\columnwidth}{!}{

\begin{tabular}{c|cccc}
\hline
NO keypoint & 1      & 3      & 6      & 10     \\ \hline
base        & 97.3\% & 97.3\% & 97.8\% & 96.9\% \\
large       & 97.6\% & 99.2\% & 98.9\% & 97.3\% \\
huge        & 98.8\% & 99.6\% & 99.2\% & 98.3\% \\
\hline
\end{tabular}

}
\caption{Effectiveness of Keypoint Numbers (1, 3, 6, 10) for Animal Re-Identification}
\label{tab:num_kp} 

\end{table}

% Table~\ref{tab:num_kp} shows the effectiveness of CKPE with different number of keypoints from 0 to 10. It indicates that our methods obtains solid improvement on different number of keypoints. It demonstrates that additional keypoints do not enhance performance. 

Table~\ref{tab:num_kp} presents the performance of Categorical Keypoint Positional Embedding (CKPE) with varying numbers of keypoints, ranging from 1 to 10. The results demonstrate that our method achieves consistent and solid improvements across different numbers of keypoints, with performance gradually stabilizing as the number of keypoints increases. Notably, the addition of more than a few keypoints does not result in significant gains, suggesting that the model has reached an optimal balance between keypoint quantity and performance. This indicates that while a small number of well-chosen keypoints can substantially improve the model’s accuracy, further increasing the number of keypoints does not provide additional benefits, and may even introduce redundancy. These findings highlight that the most critical factor in keypoint selection is the quality and relevance of the chosen keypoints, rather than their sheer quantity.

\subsection{Keypoints vs Random points}

\begin{table}[ht]
\centering
\resizebox{0.6\columnwidth}{!}{

\begin{tabular}{c|cc}
\hline
ViT     & random & keypoint \\ \hline
base  & 95.6\% & 97.3\%   \\
large & 96.2\% & 97.6\%   \\
huge  & 98.8\% & 98.8\%  \\ \hline
\end{tabular}

}
\caption{Comparison between performance of 1 keypoint vs 1 random point}
\label{tab:random_vs_kp1} 

\end{table}

\begin{table}[ht]
\centering
\resizebox{0.6\columnwidth}{!}{

\begin{tabular}{c|cc}
\hline
ViT     & random & keypoint \\ \hline
base  & 96.9\% & 98.5\%   \\
large & 98.8\% & 99.2\%   \\
huge  & 99.2\% & 99.6\%  \\ \hline
\end{tabular}

}
\caption{Comparison between performance of  3 keypoints vs 3 random points}
\label{tab:random_vs_kp3}

\end{table}

Tables~\ref{tab:random_vs_kp1} and~\ref{tab:random_vs_kp3} compare the effectiveness of keypoints against randomly selected points, using 1 and 3 points, respectively. The results clearly demonstrate that keypoints provide significant improvements over randomly chosen points, highlighting their crucial role in enhancing the performance of animal re-identification (ReID). Specifically, keypoints enable the model to focus on discriminative regions of the image, which are essential for accurately identifying and distinguishing between individuals. In contrast, random points do not provide the same level of semantic or spatial relevance, leading to suboptimal performance. These findings underscore the importance of selecting informative keypoints, as they provide valuable spatial context and semantic meaning that random points cannot replicate, ultimately playing a critical role in boosting the model’s ability to perform robustly in wildlife ReID tasks.
\section{Conclusion}
\label{sec:Conclusion}

In this work, we introduce a novel keypoint propagation mechanism, providing a potential solution for keypoint detection in wildlife datasets. We also propose a new approach for wildlife re-identification (ReID) that combines keypoint-based feature extraction with Keypoint Positional Embedding (KPE) and Categorical Keypoint Positional Embedding (CKPE). Our method achieves significant performance improvements across multiple wildlife ReID benchmarks. We demonstrate that keypoint-based embeddings, particularly when enhanced with categorical information, play a crucial role in improving model accuracy. We believe that our approach, along with the insights gained from ablation studies, will contribute to the advancement of wildlife re-identification and related ecological monitoring tasks.

{
    \small
    \bibliographystyle{ieeenat_fullname}
    \bibliography{main}
}


\end{document}